%% file: main.tex
\documentclass{article}

    \PassOptionsToPackage{numbers, compress}{natbib}

\usepackage[preprint]{neurips_2024}

\usepackage{natbib}
\usepackage{times}
\usepackage{latexsym}
\usepackage{multirow}
\usepackage{array}

\usepackage[utf8]{inputenc} 
\usepackage[T1]{fontenc}    
\usepackage{hyperref} 
\usepackage{cleveref}
\usepackage{url}            
\usepackage{booktabs}       
\usepackage{amsfonts}       
\usepackage{nicefrac}       
\usepackage{microtype}      
\usepackage[table]{xcolor}         
\usepackage{enumitem}
\usepackage{multicol}
\usepackage{multirow}
\usepackage{subcaption}
\usepackage{graphicx}
\usepackage{xspace}
\usepackage{comment}
\usepackage{adjustbox}
\usepackage[breakable]{tcolorbox}
\usepackage{fix-cm}

\input{macros}

\title{Beyond the Binary: Capturing Diverse Preferences With Reward Regularization}

\author{Vishakh Padmakumar\thanks{Equal contribution.} \\
  New York University \\
  \texttt{vishakh@nyu.edu} \\\And
  Chuanyang Jin$^*$ \\
  Johns Hopkins University \\
  \texttt{cjin33@jhu.edu} \\\And
  Hannah Rose Kirk$^*$ \\
  University of Oxford \\
  \texttt{hannah.kirk@oii.ox.ac.uk} \\\And
  He He \\
  New York University \\
  \texttt{hehe@cs.nyu.edu} \\}

\begin{document}
\maketitle
\begin{abstract}

Large language models (LLMs) are increasingly deployed via public-facing interfaces to interact with millions of users, each with diverse preferences. Despite this, preference tuning of LLMs predominantly relies on reward models trained using binary judgments where annotators select the preferred choice out of pairs of model outputs. In this work, we argue that this reliance on binary choices does not capture the broader, aggregate preferences of the target user in real-world tasks. We propose a taxonomy that identifies two dimensions of subjectivity where different users disagree on the preferred output—namely, the \textit{Plurality of Responses to Prompts}, where prompts allow for multiple correct answers, and the \textit{Indistinguishability of Responses}, where candidate outputs are paraphrases of each other. We show that reward models correlate weakly with user preferences in these cases.

As a first step to address this issue, we introduce a simple yet effective method that augments existing binary preference datasets with synthetic preference judgments to estimate potential user disagreement. Incorporating these via a margin term as a form of regularization during model training yields predictions that better align with the aggregate user preferences.
\end{abstract}

\input{1_intro}
\input{2_formulation}
\input{3_categories}

\input{4_conc}

\section*{Social Impacts Statement}
\label{sec:social_impacts}
Our work situates itself in a methodological issue of alignment fine-tuning: the overreliance on single-annotator binary judgments. We propose regularization with synthetic preferences a simple, pragmatic solution that offers a quick and easily implementable patch to improve the calibration of preference signals, especially in subjective regions of input-output space and in the absence of extensive additional human data. However, our work raises important epistemological and ethical considerations. Our assumption that LLMs can better approximate preference heterogeneity than individual annotators has mixed empirical validation \cite{argyleOut2022,mohta2023large, manning2024automated,wang2024large}. The use of synthetic annotations, while a pragmatic solution, may introduce algorithmic biases that could affect model outputs in opaque ways, misrepresenting populations of users \cite{santurkar2023whose,agnewIllusion2024}. Our demonstrated implementation adopts a utilitarian approach, simply taking a majority vote for preference aggregation. While our method is flexible to other specifications, our work does not resolve fundamental ethical questions about how to aggregate conflicting preferences or fairly represent minority viewpoints without succumbing to the tyranny of the majority. The assumption that we need to aggregate preferences may itself be flawed. In contrast to seeking a monolithic LLM with lofty ambitions to simultaneously represent the preferences of a vast user base, a promising path to pluralistic AI may come from a more granular unit of alignment via personalization or community-specific steering and cultural fine-tuning \citep{sorensenRoadmap2024}. We present this work not as a definitive solution, but as a contribution to the ongoing discourse on responsible, safe, and inclusive AI development, emphasizing the need for continued interdisciplinary research to address the sociotechnical and normative challenges central to technical alignment methodologies.

\section*{Acknowledgment}
This work is supported by Open Philanthropy, AWS
AI, and the National Science Foundation under Grant No. 1922658 and Grant No. IIS-2340345.

\bibliographystyle{unsrtnat}
\bibliography{custom, references}

\appendix
\input{appendix}

\end{document}

%% file: macros.tex
\newcommand{\PreserveBackslash}[1]{\let\temp=\\#1\let\\=\temp}
\newcolumntype{C}[1]{>{\PreserveBackslash\centering}p{#1}}
\newcolumntype{R}[1]{>{\PreserveBackslash\raggedleft}p{#1}}
\newcolumntype{L}[1]{>{\PreserveBackslash\raggedright}p{#1}}

\definecolor{red}{RGB}{255, 0, 0}
\definecolor{blue}{RGB}{135, 206, 250}
\definecolor{green}{RGB}{205, 255, 204}

\newcommand{\ie}{i.e.\xspace}

%% file: 1_intro.tex
\section{Introduction}
\label{sec:intro}


A ubiquitous step in the training of contemporary large language models (LLMs) is aligning their output with human preferences \citep{ouyang2022training, bai2022training, touvron2023llama, achiam2023gpt, feng2024far}. This process involves collecting human preference judgments over model outputs, which serve as the reward signal for either reinforcement learning with human feedback \citep{ouyang2022training, stiennon2020learning} or various direct alignment algorithms \citep{rafailov2024direct, meng2024simpo}. Conventionally, a single human annotator provides a binary judgment for a pair of model outputs \citep{lambertHistory2023, kirk2023past}.

This ``preferentist'' approach to alignment has been criticized \citep{gabrielArtificial2020,tasioulasArtificial2022, zhi2024beyond}, pitted against alternatives such as deliberative procedures that establish normative standards among diverse stakeholders \citep{, changMeta2024, bergmanSTELA2024} or principle-based frameworks that seek higher-order rules or ``constitutions'' to guide AI behavior \citep{bai2022constitutional, glaese2022improving, huang2024collective}. However, preference judgments remain central to LLM post-training due to their cognitive and cost \textit{efficiency}---it is far less demanding for annotators to choose between two outputs than to articulate an ideal demonstration---and surprising \textit{efficacy} in steering LLMs towards being more friendly, helpful, and adept at interpreting user intent \citep{ouyang2022training, bai2022training, metaaiIntroducing2024, jin2024mmtom}. So, even if preference fine-tuning is not the optimal solution, it is likely to remain a key component in LLM development for the foreseeable future, especially in industry settings.


However, despite its ubiquity and early success, this dominant methodology for preference-tuning LLMs is now at odds with the reality of their deployment in society. Accessible via public interfaces, LLMs have amassed hundreds of millions of users \citep{rothChatGPT2024}, who have diverse preferences, needs, and linguistic backgrounds. The single binary choice between outputs 
becomes an unreliable signal 
for this vast user base, particularly in subjective domains like safety \citep{thoppilan2022lamda, aroyoDICES2023} or toxicity \citep{korbak2023pretraining}, and when judging the overall quality of the output \citep{stiennon2020learning} in absence of detailed guidelines for what counts as high-quality language \citep{kirk2023empty, gururanganWhose2022}. 


In this paper, we address this inherent gap in development-deployment preference formulation. We argue that if preference tuning continues to be a crucial component in modern LLM training, steps must be taken to calibrate the preference signal so that it better translates from a single-annotator development setting to a multi-user deployment environment. We present our empirical argument in two phases:

\textbf{The ``break it'' phase} demonstrates the need for more calibrated reward estimation through a detailed error analysis of existing reward models. We hypothesize that over-reliance on single-annotator judgments leads to noisy and poorly calibrated preference signals, especially on subjective examples where different annotators would disagree on the preferred choice. We present a qualitative taxonomy of reward model failures in such cases (\Cref{sec:subj_defn}) and validate that reward model predictions on these examples correlate more poorly with human judgments (\Cref{sec:annotation}. 

\textbf{The ``fix it'' phase} tests 
a simple yet effective modification to the reward model training objective. 
Based on recent work \cite{touvron2023llama}, we introduce a margin term in the objective function of reward modeling that is scaled to reflect the degree of disagreement of preference judgments in a hypothetical population of users. Our experiments show that these regularized model predictions better align with human preferences (\Cref{sec:smoothing}). 


Our contributions add to a body of work seeking alignment of LLMs that can better account for natural distributions of human preferences across global populations \citep{kirkPRISM2024, jin2023cultural, zhang2024diverging}. As a resolution to preference disagreements, some prior work has targeted personalization \citep{jang2023personalized, liPersonalized2024, poddar2024personalizing}, but our approach is more similar to those seeking more principled approaches to aggregate preference estimation \citep{rame2024rewarded}, for example, those drawing insight from social choice theory \citep{bakker2022fine, chakraborty2024maxmin, conitzerSocial2024}, seeking distributional learning \citep{siththaranjanDistributional2023, liAligning2024} or uncertainty modeling \citep{lou2024uncertainty}. Our margin-based method can be seen as adopting a utilitarian approach to preference maximization with uniform weightings, but the framework can also accommodate other normative considerations of distributive justice in AI alignment.


\begin{figure}
    \centering
    \includegraphics[width=\linewidth]{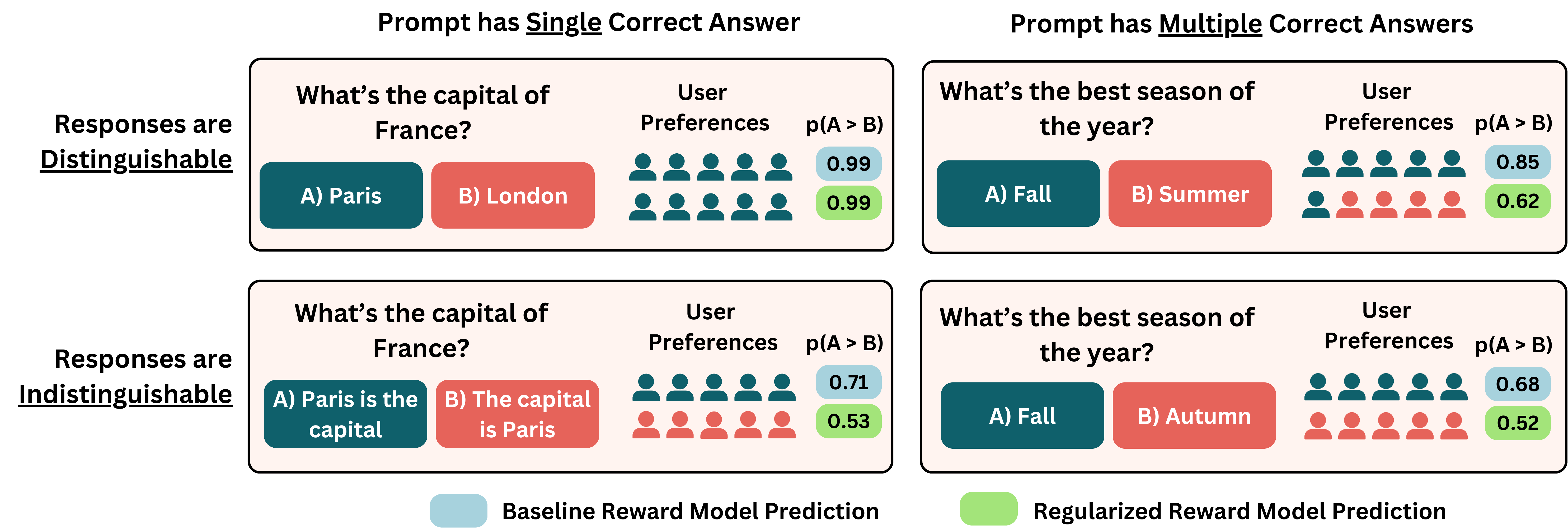}
    \caption{Two dimensions of subjectivity along which reward model predictions correlate poorly with the diverse preferences of a population of users---when the prompt allows for multiple correct answers and when the two responses are indistinguishable from one another (\Cref{sec:categorization}). Regularizing 
    the training objective with a margin term 
    (\Cref{sec:smoothing}) leads to predictions that better correlate with these preferences. 
    \vspace{-0.5cm}
    }
    \label{fig:fig_1}
\end{figure}


Our method represents a simple impactful adjustment that is compatible with existing dominant paradigms of preference fine-tuning, though it is not a complete solution. By further refining our model and data methodologies for preference alignment, we can work towards AI systems that better accommodate the diversity of subjective human experiences.

\vspace{-0.25cm}

%% file: 2_formulation.tex
\section{Problem Formulation}
\label{sec:formulation}

The canonical method to train a reward model, $r_\theta$, involves collecting a human preference judgment, $j$, that indicates which of two candidate outputs $y \in \{y_0, y_1\}$ is preferred for a given input prompt $x$, resulting in dataset $\mathrm{D}$. The model parameters are optimized by minimizing the loss function:

\begin{equation}
\textrm{loss}(r_\theta) = - \mathrm{E}_{(x, y_0, y_1, j) \sim \mathrm{D}} [\log (\sigma(r_\theta(x, y_{j}) - r_\theta(x, y_{1-{j}}))) ],    
\label{eqn:original}
\end{equation}

where $\sigma$ denotes the sigmoid function and $j \in \{0, 1\}$.

The reward model parameters are trained to maximize the difference in score assigned to $y_{j}$ 
and $y_{1-j}$
based on the preference judgment, $j$.
This is effective when $j$ is unanimously the `better' output, as $j$ would likely be consistent with the preferences of any future user interacting with the model. 

However, we argue that in practical applications of LLMs, the models are frequently required to make decisions on examples where there is legitimate disagreement among users. Consider a scenario where a set of $n$ users each provide a judgment, $j_{1 \dots n}$, on the same example. The intended behavior of the model is to predict the aggregate preference of the population, specifically the fraction of users that prefer option $y_0$ over $y_1$. We write this as:
\begin{equation}
p^*_{01} = \frac{\sum_{i=1}^n \mathbf{I}(j_i == 0)}{n}
\label{eqn:aggregate}
\end{equation}
where $\mathbf{I}(\cdot)$ is and indicator function that equals $1$ if user $i$ prefers option $y_0$, and $0$ otherwise.

However, this true preference $p^*_{01}$ is not accessible since we do not have prior knowledge about future user preferences. This can lead to poor reward model performance when $p^*_{01}$ deviates from the binary judgment $j$ in \Cref{eqn:original}. In \Cref{sec:categorization}, we outline two dimensions along which subjective examples can be identified (\Cref{sec:subj_defn}) and empirically demonstrate that the correlation between reward model predictions and the aggregate user preferences—our estimate of $p^*_{01}$—is weaker for these subjective examples (\Cref{sec:annotation}). 


\paragraph{Can we use synthetic preference judgments to make more aligned predictions?}

As part of our initial efforts to mitigate this effect, we propose a simple modification to the reward modeling objective. Here we take inspiration from \citet{touvron2023llama} who collect and incorporate the \emph{extent} of user preferences between pairs of model outputs, \ie \emph{significantly} better vs \emph{slightly} better, via a margin term scaled from $0$ to $1$. We introduce a similar margin term that scales proportional to the estimated disagreement on the choice between outputs $y_0$ and $y_1$ given prompt $x$---the margin value should be high when the choice between the two options is unanimous and vice versa. 
Our margin term converts binary preference judgments from different annotators into a cardinal measure of the \emph{strength} of preference of the group as a whole. A gold standard implementation of the margin would be to extend to the preference data collection process to obtain multiple judgments for each pair of outputs from different annotators. However, the prohibitive cost of re-annotating existing datasets makes us favor a silver version that implements the margin using synthetic judgments from an LLM. Given an input prompt and a two candidate outputs $(x, y_0, y_1)$, we sample $n$ synthetic judgments $j'_{i=1 \dots n} \in \{0, 1\}$ for a binary preference between the two outputs.
We convert these into a margin term $m_{x, y_0, y_1} \in 
[0, 1]$ as:
\begin{equation}
m_{x, y_0, y_1} = {\Big(\Big|\sum_{i=1}^{n} j'_i - \frac{n}{2}\Big|\Big)} \; \Big/ \; {\frac{n}{2}}
\label{eqn:margin}
\end{equation}

Here $m=1$ if all of the synthetic judgments are either $0$ or $1$, \ie a unanimous preference in a population of annotators, and $m=0$ if half the of the synthetic judgments are 0 and 1 each, \ie a highly contested preference.\footnote{We scale the margin between $0$ and $1$ following empirical experiments on the range of the margin in \citet{touvron2023llama}.}

This margin term is incorporated into the loss objective as follows:
\begin{equation}
\textrm{loss}(r_\theta) = - \mathrm{E}_{(x, y_0, y_1, j) \sim \mathrm{D}} [\log (\sigma(r_\theta(x, y_{j}) - r_\theta(x, y_{j_{1-j}}) - m_{x, y_0, y_1}))]
\label{eqn:obj_smoothing}
\end{equation}

We test the effect of this intervention in \Cref{sec:smoothing} and demonstrate how this improves the correlation of reward model predictions to user preferences in subjective examples.


%% file: 3_categories.tex
\section{Breaking Down Reward Model Performance by Category}
\label{sec:categorization}

In this section, we sent a categorization of examples according to the degree of disagreement in aggregate preference among annotators (\Cref{sec:subj_defn}) and demonstrate that the performance of reward models drops in those categories with high levels of disagreement (\Cref{sec:annotation}). We then show how the performance improves by incorporating margin-based regularization into the model training pipeline (\Cref{sec:smoothing}).

\subsection{Categories of Subjectivity}
\label{sec:subj_defn}
In current frameworks for training language models to learn human preferences, such as Reinforcement Learning from Human Feedback (RLHF), the predominant method of preference annotation relies on binary judgments between pairs of model-generated outputs. However, human preferences are inherently subjective and exhibit significant inter-individual variability. Hence, singular binary annotations may not sufficiently capture the range of opinions. 

To better accommodate this variability in human judgment, we first propose to classify examples into two primary categories based on the nature of their prompts: (i) prompts that possess an objective, \emph{single correct} answer, and (ii) prompts that are inherently subjective, characterized by admitting \emph{multiple correct} responses.

Moreover, within existing datasets, we observe many instances where the two model-generated responses are paraphrases of each other. This can occur in responses to both objective and subjective prompts. Therefore, we suggest a secondary dimension of categorization based on the distinguishability of the two model-generated candidates: (i) responses that are \emph{distinguishable}, and (ii) responses that are \emph{indistinguishable} paraphrases.

\Cref{fig:fig_1} details illustrative examples from each of the aforementioned categories. We now evaluate whether this categorization corresponds to a decrease in the correlation between reward model predictions and user preference judgments.


\subsection{Evaluating Reward Model Performance} 
\label{sec:annotation}

\paragraph{Experimental Setup}
We use a trained {DeBERTa-V3} as the reward model for our experiments.\footnote{We selected this reward model, released as part of the \href{https://huggingface.co/OpenAssistant/reward-model-deberta-v3-large-v2}{Open Assistant project}, for our experiments due to its publicly available training data and the fact that it was the most 
\href{https://huggingface.co/models?sort=downloads&search=reward+model}{frequently downloaded} reward model on Huggingface Hub at the time of our experiments.}
Our goal is to evaluate its performance across examples in each of the aforementioned categories: Multiple/Single Correct Answer; Distinguishable/Indistinguishable Responses. We construct our test set of 150 examples by randomly selecting 25 examples from each of six datasets. We use four in-domain (ID) datasets that were used to train the reward model
---\textsc{WebGPT} \citep{nakano2021webgpt}, HH-RLHF \citep{bai2022training}, Open AI Summarize \citep{stiennon2020learning}, and \textsc{InstructGPT-J} \citep{alex_havrilla_2023}. The remaining two datasets, \textsc{Prism} \citep{kirkPRISM2024} and \textsc{UltraFeedback} \citep{cui2023ultrafeedback}, serve as out-of-domain (OOD) sets used for testing purposes. 

We recruit three human annotators to label the 150 example pairs manually. We first put the examples into categories along both aforementioned dimensions  
via a majority vote. To estimate the aggregate preference of future users (\Cref{eqn:aggregate}), 
our annotators are instructed as follows: ``If you asked 10 people, how many would prefer answer A and how many would prefer answer B?'' This allows them to provide a judgment on a scale ranging from 10-0 to 0-10, and we average these responses to obtain a consensus.
Notably, this approach enables participants to first predict the believed subjectivity—even if they personally prefer answer A, they might project that only 2 out of 10 people would share their preference, thus providing insight into population-level preferences. Empirical evidence suggests that this approach yields more reliable results than direct population votes asking each annotator for their preference between answer A and answer B. We evaluate the performance of the reward model through the Pearson correlation (\Cref{tab:results}) and L1 loss (\Cref{tab:results_l1}) between the collected human preference and this \emph{Baseline} reward model (RM) predictions. 


\paragraph{Results} From \Cref{tab:results} and \Cref{tab:results_l1}, we see that the model performs significantly worse on subjective examples, indicating its limitations in capturing human preferences where \emph{multiple correct} responses exist. Moreover, the model demonstrates notably poorer performance on the OOD datasets, suggesting that traditional reward modeling approaches may not adequately generalize across different data domains. Additionally, we find that the baseline reward model performs similarly (\Cref{tab:results}), or slightly better (\Cref{tab:results_l1}) on indistinguishable than distinguishable responses, indicating slightly more robust behavior along this axis of subjectivity. Our findings align with contemporary work that investigate annotator disagreement along different dimensions to also conclude that current reward models, which are trained with the assumption that this disagreement is noise, inadequately represent these diverse preferences \citep{zhang2024diverging}.



\begin{table*}[t] 
\setlength{\tabcolsep}{2pt}
\setlength{\arrayrulewidth}{0pt}
\fontsize{8.5}{11}\selectfont

\begin{adjustbox}{valign=b,raise=\baselineskip}
\begin{tabular}{R{3cm}C{1cm}C{1.3cm}C{0.9cm}}
\toprule
 \textbf{Dataset (N)} & \textbf{Baseline RM} & \textbf{Regularized RM} & \textbf{$\Delta$} \\ 
\midrule
\textbf{\textsc{WebGPT} (25)} & \cellcolor[rgb]{0.612,0.843,0.729}0.677 & \cellcolor[rgb]{0.647,0.859,0.757}0.637 & \cellcolor[rgb]{0.957,0.8,0.8}-0.040 \\
\textbf{\textsc{HH-RLHF} (25)} & \cellcolor[rgb]{0.341,0.733,0.541}0.974 & \cellcolor[rgb]{0.341,0.733,0.541}0.974 & 0.000 \\
\textbf{\textsc{OaiSummarize} (25)} & \cellcolor[rgb]{0.804,0.922,0.863}0.468 & \cellcolor[rgb]{0.698,0.878,0.788}0.585 & \cellcolor[rgb]{1,0.902,0.624}0.117 \\
\textbf{\textsc{InstructGPT-J} (25)} & \cellcolor[rgb]{0.706,0.882,0.796}0.573 & \cellcolor[rgb]{0.733,0.894,0.816}0.544 & \cellcolor[rgb]{0.969,0.855,0.855}-0.029 \\
\midrule
\textbf{All ID (100)} & \cellcolor[rgb]{0.561,0.824,0.694}0.734 & \cellcolor[rgb]{0.549,0.82,0.686}0.749 & \cellcolor[rgb]{1,0.988,0.953}0.015 \\ \midrule \midrule
\textbf{\textsc{Prism} (25)} & \cellcolor[rgb]{0.8,0.922,0.859}0.472 & \cellcolor[rgb]{0.631,0.851,0.741}0.658 & \cellcolor[rgb]{1,0.839,0.4}0.186 \\
\textbf{\textsc{UltraFeedback} (25)} & 0.248 & \cellcolor[rgb]{0.976,0.992,0.984}0.277 & \cellcolor[rgb]{1,0.976,0.91}0.029 \\
\midrule
\textbf{All OOD (50)} & \cellcolor[rgb]{0.886,0.953,0.922}0.377 & \cellcolor[rgb]{0.784,0.914,0.851}0.488 & \cellcolor[rgb]{1,0.906,0.643}0.111 \\ 
\bottomrule
\end{tabular}
\end{adjustbox}%
\hfill
\begin{adjustbox}{valign=b,raise=\baselineskip}
\renewcommand{\arraystretch}{1.025}
\begin{tabular}{R{3.45cm}C{1cm}C{1.3cm}C{0.9cm}}
\toprule
\textbf{Category (N)} & \textbf{Baseline RM} & \textbf{Regularized RM} & \textbf{$\Delta$} \\ 
\midrule
\textbf{Multiple Correct (106)} & \cellcolor[rgb]{0.71,0.882,0.796}0.572 & \cellcolor[rgb]{0.659,0.863,0.765}0.626 & \cellcolor[rgb]{1,0.957,0.827}0.054 \\
\textbf{Single Correct (44)} & \cellcolor[rgb]{0.498,0.8,0.651}0.802 & \cellcolor[rgb]{0.502,0.8,0.655}0.799 & \cellcolor[rgb]{0.996,0.984,0.984}-0.003 \\ \midrule \midrule
\textbf{Multiple Correct ID (66)} & \cellcolor[rgb]{0.6,0.839,0.722}0.691 & \cellcolor[rgb]{0.576,0.831,0.706}0.715 & \cellcolor[rgb]{1,0.98,0.925}0.024 \\
\textbf{Single Correct ID (40)} & \cellcolor[rgb]{0.471,0.788,0.631}0.835 & \cellcolor[rgb]{0.478,0.792,0.639}0.824 & \cellcolor[rgb]{0.984,0.941,0.941}-0.011 \\
\textbf{Multiple Correct OOD (34)} & 0.251 & \cellcolor[rgb]{0.878,0.953,0.918}0.384 & \cellcolor[rgb]{1,0.886,0.573}0.133 \\
\textbf{Single Correct OOD (10)} & \cellcolor[rgb]{0.596,0.839,0.722}0.694 & \cellcolor[rgb]{0.533,0.812,0.675}0.766 & \cellcolor[rgb]{1,0.941,0.769}0.072 \\ \midrule \midrule
\textbf{Distinguishable (124)} & \cellcolor[rgb]{0.647,0.859,0.753}0.640 & \cellcolor[rgb]{0.604,0.839,0.725}0.686 & \cellcolor[rgb]{1,0.961,0.855}0.046 \\
\textbf{Indistinguishable (26)} & \cellcolor[rgb]{0.655,0.863,0.761}0.631 & \cellcolor[rgb]{0.635,0.855,0.745}0.652 & \cellcolor[rgb]{1,0.984,0.933}0.021 \\ \bottomrule
\end{tabular}
\end{adjustbox}

\caption{Pearson correlation of predictions from the baseline reward model (RM) (\Cref{sec:annotation}) and regularized RM (\Cref{sec:smoothing}) to user preferences. \emph{Regularization} improves performance on both OOD datasets as well as subjective examples with multiple valid answers. Cells in green show absolute performance, while $\Delta$ shows (Regularized RM - Baseline RM).}
\label{tab:results}
\end{table*}
\vspace{10mm}

\begin{table*}[t] 
\setlength{\tabcolsep}{2pt}
\setlength{\arrayrulewidth}{0pt}
\fontsize{8.5}{11}\selectfont

\begin{adjustbox}{valign=b,raise=\baselineskip}
\begin{tabular}{R{3cm}C{1.5cm}C{1.5cm}}
\toprule
\textbf{Dataset (N)} & \textbf{Baseline RM} & \textbf{Regularized RM} \\ \midrule
\textbf{\textsc{WebGPT} (25)} & 0.171 & \textbf{0.159}  \\
\textbf{\textsc{HH-RLHF} (25)} & 0.114 & \textbf{0.108} \\
\textbf{\textsc{OaiSummarize} (25)} & 0.274 & \textbf{0.207} \\
\textbf{\textsc{InstructGPT-J} (25)} & \textbf{0.315} & 0.319 \\
\midrule
\textbf{All ID (100)} & 0.218 & \textbf{0.198} \\ \midrule \midrule
\textbf{\textsc{Prism} (25)} & 0.275 & \textbf{0.205} \\
\textbf{\textsc{UltraFeedback} (25)} & \textbf{0.255} & 0.270 \\
\midrule
\textbf{All OOD (50)} & 0.265 & \textbf{0.238} \\ 
\bottomrule
\end{tabular}
\end{adjustbox}%
\hfill
\begin{adjustbox}{valign=b,raise=\baselineskip}
\renewcommand{\arraystretch}{1.025}
\begin{tabular}{R{3.45cm}C{1.5cm}C{1.5cm}}
\toprule
\textbf{Category (N)} & \textbf{Baseline RM} & \textbf{Regularized RM} \\ 
\midrule
\textbf{Multiple Correct (106)} & 0.263 & \textbf{0.234} \\
\textbf{Single Correct (44)} & 0.163 & \textbf{0.158} \\ \midrule \midrule
\textbf{Multiple Correct ID (66)} & 0.252 & \textbf{0.227} \\
\textbf{Single Correct ID (40)} & 0.152 & \textbf{0.142} \\
\textbf{Multiple Correct OOD (34)} & 0.281 & \textbf{0.244} \\
\textbf{Single Correct OOD (10)} & \textbf{0.202} & 0.211 \\ \midrule \midrule
\textbf{Distinguishable (124)} & 0.244 & \textbf{0.219} \\
\textbf{Indistinguishable (26)} & 0.176 & \textbf{0.160} \\ \bottomrule
\end{tabular}
\end{adjustbox}

\caption{Average L1 loss (absolute difference) between the aggregate user preferences (normalized to a fraction from $0$ to $1$) and the model predictions from the baseline RM (\Cref{sec:annotation}) and regularized RM (\Cref{sec:smoothing}). Bold values deviate less from the user preferences. \emph{Regularization} helps RMs better capture user preferences, particularly on average across all ID and OOD datasets as well as subjective examples with multiple valid answers.}
\label{tab:results_l1}
\end{table*}

\subsection{Regularization With Synthetic Preferences to Improve Performance}
\label{sec:smoothing}

Having observed the decline in reward model performance on subjective examples
, we test if training models with our proposed margin-based intervention 
results in better generalization to these examples. To do so, we first obtain $10$ synthetic annotations, $j'_{i=1 \dots 10}$, for each example in the $4$ training datasets (\Cref{sec:annotation}) using the Llama-3 70B Instruct model \cite{dubey2024llama}.\footnote{We select the meta-llama/Meta-Llama-3-70B-Instruct via Huggingface \cite{wolf2020transformers} as it was the highest performing \emph{open-weight} model on the Stanford HELM benchmark \cite{liang2022holistic} at the time we conducted our experiments.} We generate synthetic annotations with nucleus sampling setting the top-p parameter to $0.9$ to mimic the variability of a group of annotators.\footnote{We provide the exact prompt used in \Cref{sec:prompts}.} These are used to construct the margin for each training example as detailed in \Cref{eqn:margin}.
We then train the Deberta-V3 reward model using the objective in \Cref{eqn:obj_smoothing} \footnote{To clarify, we fine-tune a pre-trained \emph{Deberta-V3} model on the same training data as the baseline model from \Cref{sec:annotation} using our modified objective. We run a hyperparameter sweep of learning rates from $1e-04$ to $1e-06$ and select the best-performing model on the validation data.} Finally, we report the Pearson correlation and L1 loss between the model predictions and the collected human annotations on the test set in \Cref{tab:results}. We observe that our method results in better performance on the \emph{multiple correct} subsets, both ID and OOD, without a degradation in the corresponding \emph{single correct} subsets. Our method also results in an improvement in performance scores on examples where the two outputs are \emph{indistinguishable}, without costing performance on the \emph{distinguishable} subset.

%% file: 4_conc.tex
\section{Conclusion}
In this work, we contend that reward models trained on binary preferences struggle to generalize to subjective examples where different users may disagree on the preferred option.
By categorizing examples along two dimensions of subjectivity—when the prompt allows for multiple correct answers and when the candidate outputs are paraphrases of each other—we demonstrate that reward model predictions exhibit a weaker correlation with human preferences in these cases. 
To address this, we propose a margin-based regularization technique that mitigates this issue using synthetic annotations from an LLM to improve prediction quality in subjective scenarios. We re-iterate the scope of this solution as a step to improve reward model performance as they are currently deployed, a technique to be used when recollecting large-scale preference data is not possible financially. We detail further social concerns that arise below. 


%% file: appendix.tex
\section{Prompts for Obtaining Synthetic Preference Judgments}
\label{sec:prompts}

To gather synthetic preference judgments from Llama-3 in \Cref{sec:smoothing}, we use the following prompt:

\begin{tcolorbox}[
    colframe=green!70!black,
    colback=green!10,
    coltitle=white,
    fonttitle=\bfseries,
    title=Prompt for Obtaining Synthetic Preference Judgments,
    colbacktitle=green!70!black
]
I am going to give you a prompt and two answers.

The goal is to identify the better answer.

If the prompt is a question, we want the factually correct answer.

In a conversation, we want helpful replies that do not cause harm.

The output format should be either 'Choice: A' or 'Choice: B' based on the selected answer and nothing else.

Prompt: \{prompt\}

Answer A: \{answer\_0\}

Answer B: \{answer\_1\}
\end{tcolorbox}

\section{Annotator Guidelines}
\label{sec:annotation_guidelines}

We outline the guidelines provided to annotators for conducting the manual annotations in \Cref{sec:annotation}:

\vspace{3mm}

\begin{tcolorbox}[
    colframe=blue!70!black,
    colback=blue!10,
    coltitle=white,
    fonttitle=\bfseries,
    title=Task 1: Subjectivity,
    colbacktitle=blue!70!black
]
Consider the following prompt:
\{prompt\}

Is there a single factually, correct response to this prompt?

Options:

A) There is only one single correct choice to this prompt.

B) There are multiple possible correct answers to this prompt
\end{tcolorbox}

\vspace{3mm}

\begin{tcolorbox}[
    colframe=blue!70!black,
    colback=blue!10,
    coltitle=white,
    fonttitle=\bfseries,
    title=Task 2: Distinguishability,
    colbacktitle=blue!70!black
]
Consider the following two candidate model outputs 
in response to the prompt: \{prompt\} 

Answer A: \{answer\_0\} 

Answer B: \{answer\_1\}

Can you clearly tell the two responses apart? Are they significantly different from one another?

Options: 

A) Both choices are essentially the same content, just with minor stylistic variations.

B) The two choices are clearly different from each other and distinguishable.
\end{tcolorbox}

\vspace{3mm}
\begin{tcolorbox}[
    colframe=blue!70!black,
    colback=blue!10,
    coltitle=white,
    fonttitle=\bfseries,
    title=Task 3: User Preference Estimation,
    colbacktitle=blue!70!black
]
If you asked 10 different people, how many would prefer answer A and how many would prefer answer B?

1) 10 A - 0 B

2) 9 A - 1 B

3) 8 A - 2 B

4) 7 A - 3 B

5) 6 A - 4 B

6) 5 A - 5 B

7) 4 A - 6 B

8) 3 A - 7 B

9) 2 A - 8 B

10) 1 A - 9 B

11) 10 A - 0 B
\end{tcolorbox}
